\documentclass[runningheads]{llncs}
\usepackage{graphicx}
\usepackage{pifont}
\newcommand{\cmark}{\ding{51}}%
\newcommand{\xmark}{\ding{55}}%
\usepackage{caption}
\usepackage{graphicx,subcaption}

\usepackage{tikz}
\usetikzlibrary{arrows.meta}
\usepackage{comment}
\usepackage{amsmath,amssymb} 
\usepackage{color}

\usepackage[accsupp]{axessibility}  

\usepackage[utf8]{inputenc}
\usepackage{pgfplots}
\DeclareUnicodeCharacter{2212}{−}
\usepgfplotslibrary{groupplots,dateplot}
\usetikzlibrary{patterns,shapes.arrows}
\pgfplotsset{width = 6cm}
\usepackage{xcolor}
\usepackage{adjustbox}
\usepackage{multirow}
\usepackage{tabularray}
\UseTblrLibrary{booktabs}
\usepackage{float}
\usetikzlibrary{positioning}
\begin{document}
\pagestyle{headings}
\mainmatter
\def\ECCVSubNumber{6655}  

\title{S3C: Self-Supervised Stochastic Classifiers for Few-Shot Class-Incremental Learning} 

\titlerunning{S3C: Self-Supervised Stochastic Classifiers for FSCIL}
%
\author{Jayateja Kalla \and
Soma Biswas}
\authorrunning{J. Kalla et al.}
%
\institute{Department of Electrical Engineering, \\Indian Institute of Science, Bangalore, India. \\
\email{\{jayatejak, somabiswas\}@iisc.ac.in}}
\maketitle
\begin{abstract}
Few-shot class-incremental learning (FSCIL) aims to learn progressively about new classes with very few labeled samples, without forgetting the knowledge of already learnt classes. 
FSCIL suffers from two major challenges: (i) \textit{over-fitting} on the new classes due to limited amount of data, (ii) \textit{catastrophically forgetting} about the old classes due to unavailability of data from these classes in the incremental stages. In this work, we propose a self-supervised stochastic classifier (S3C)\footnote{\footnotesize{code: \color{blue}\url{https://github.com/JAYATEJAK/S3C}\color{black}}} to counter both these challenges in FSCIL. 
The stochasticity of the classifier weights (or class prototypes) not only mitigates the adverse effect of absence of large number of samples of the new classes, but also the absence of samples from previously learnt classes during the incremental steps. This is complemented by the self-supervision component, which helps to learn features from the base classes which generalize well to unseen classes that are encountered in future, thus reducing catastrophic forgetting. Extensive evaluation on three benchmark datasets using multiple evaluation metrics show the effectiveness of the proposed framework. We also experiment on two additional realistic scenarios of FSCIL, namely where the number of annotated data available for each of the new classes can be different, and also where the number of base classes is much lesser, and show that the proposed S3C performs significantly better than the state-of-the-art for all these challenging scenarios.
\end{abstract}

\section{Introduction}
In recent years, Deep Neural Networks (DNN) have shown significant performance improvement on various computer vision applications \cite{krizhevsky2012imagenet,noh2015learning,ouyang2016deepid}. 
Usually, the DNN models require enormous amount of annotated data from all the classes of interest to be available for training. 
In real-world, since data from different classes may become available at different instants of time, we want the model to learn about the new classes incrementally without forgetting about the old classes, which is precisely the task addressed in  Class-Incremental Learning (CIL).
CIL approaches are very useful and practical, not only because it is computationally expensive and time-consuming to retrain the model from scratch, but also data from the previous classes may not be available due to storage and privacy issues.

Since collecting large number of annotated data from all the new classes is also very difficult, recently, the more challenging but realistic few-shot class-incremental learning (FSCIL) is gaining increasing attention, where the new classes have few labeled samples per class~\cite{tao2020few}.
In FSCIL, a model is first learnt using a set of base classes with large number of labeled examples per class.
At each incremental step (task), the model has access to a few labeled samples of the new classes and a single prototype for each of the previously learnt classes. 
The goal is to learn a unified classifier to recognize the old as well as the new classes, without having access to any task labels.
This helps the model to quickly learn about the new classes without requiring to collect and annotate large amounts of data for the new classes. 
FSCIL faces two major challenges, namely overfitting due to limited samples for the new classes, and catastrophic forgetting of the already learnt classes due to absence of old classes data at the incremental steps.

In this work, we propose a novel framework, S3C (\textbf{S}elf-\textbf{S}upervised \textbf{S}tochastic \textbf{C}lassifier) to simultaneously address both these challenges in the FSCIL setting.
Unlike the standard classifiers, stochastic classifiers (SC) are represented by weight distributions, i.e. a mean and variance vector~\cite{lu2020stochastic}. 
Thus, each classifier weight sampled from this distribution is expected to correctly classify  the input samples.  
We show for the first time, that SC learnt for both the base and new classes can significantly reduce the over-fitting problem on the new classes for FSCIL task. 
It can also arrest the catastrophic forgetting of the previously learnt classes to a certain extent.
As is common in most FSCIL approaches \cite{cec,mazumder2021few}, we propose to freeze the feature extractor and learn only the SC at each incremental step. In order to compute features from the base classes which generalize to unseen classes, inspired by recent works \cite{selfsup,zhu2021prototype}, we use self-supervision along with SC giving our final S3C framework. As expected, this helps to significantly mitigate the effect of catastrophic forgetting, while at the same time retaining the advantage on the new classes.
To this end, our contributions are as follows:
\begin{enumerate}
    \item We propose a novel framework, termed S3C (Self-Supervised Stochastic Classifier) to address the FSCIL task. 
    \item We show that  stochastic classifiers can help to significantly reduce over-fitting on the new classes with limited amount of data for FSCIL.
    \item We also show that self-supervision with stochastic classifier can be used to better retain the information of the base classes, without hindering the enhanced performance of the stochastic classifiers for the new classes.
    \item We set the new state-of-the-art for three benchmark datasets, namely CIFAR100 \cite{cifar}, CUB200 \cite{cub} and miniImageNet \cite{cec}.
    \item We also propose and evaluate on two additional, realistic FSCIL settings, namely FSCIL-im (FSCIL-imbalanced) - where the new classes may have different number of samples/class and (ii) FSCIL-lb (FSCIL-less base) - where there are less number of base classes, which further justifies the effectiveness of the proposed S3C framework.
 
\end{enumerate}
\section{Related Works}
Here, we provide some pointers to the related work in literature. \\
\textbf{Class-Incremental Learning (CIL): } 
The goal of CIL is to learn new classes progressively without any task information. Due to plenty of annotated new class data, mitigating catastrophic forgetting is a challenging problem. LwF~\cite{li2017learninglwf} proposed to use knowledge distillation~\cite{hinton2015distilling} to alleviate catastrophic forgetting. iCaRL~\cite{rebuffi2017icarl} showed that nearest classifier mean (NCM) using old class exemplars can generate robust classifiers for CIL. EEIL~\cite{EEIL} used knowledge distillation to remember old classes and cross-entropy to learn new classes in an end-to-end training. UCIR~\cite{ucir} proposed cosine-based classifiers and used feature-space distillation and inter-class separation margin loss to mitigate catastrophic forgetting. Several state-of-art-works~\cite{wu2019large,zhao2020maintaining,belouadah2019il2m,douillard2020podnet,belouadah2020scail} proposed different techniques to address the class imbalance problem in CIL like rescaling scores or balanced finetuning of classifiers, etc. 
Some of the recent works~\cite{zhu2021prototype,yu2020semanticsdc,zhu2021classdual} have focused on non-exemplar based methods, with no access to exemplars from the old classes. \\ \\
\textbf{Few-Shot Class-Incremental Learning (FSCIL): }Recently, there has been a significant focus on the more realistic and challenging FSCIL task, where very few samples per class are available for training at each incremental task. 
Tao {\em et al.}~\cite{tao2020few} proposed this protocol and used neural network gas architecture to preserve the feature topologies of the base and new classes. 
Mazumder {\em et al.}~\cite{mazumder2021few} proposed to identify unimportant parameters in the model based on their magnitudes and learn only these parameters during the incremental tasks. 
The works proposed in~\cite{cheraghian2021semantic,vector_quantization,akyurek2021subspacereg,cheraghian2021mixture,zhu2021selfpromoted,lee2021fewindependentmachanisms,shi2021overcoming} focus on learning robust manifolds by regularizing feature space representations.
The works in~\cite{dong2021fewknowledgeretentiongraph,tan2021graphfewshot,cec} used graph-based networks for old classes' knowledge retention. 
Recently, CEC~\cite{cec} proposed a meta-learning strategy and achieved state-of-art results for the FSCIL setting. \\ \\
\textbf{Self-Supervised Learning (SSL): } SSL uses predefined pretext tasks to learn features from unlabeled data. 
Different pretext tasks have been proposed like image rotations~\cite{komodakis2018unsupervisedimagerotations}, image colourization~\cite{larsson2016learningimagecolorization}, clustering~\cite{caron2018deepimageclustering}, and solving jigsaw puzzles from image patch permutations~\cite{noroozi2016unsupervisedjigsaw}. 
These features can notably improve the performance of downstream tasks like few-shot learning~\cite{gidaris2019boostingselffewshot}, semi-supervised learning~\cite{zhai2019s4lselfsemisupervised}, to improve the model robustness~\cite{hendrycks2019usingselfrobustness}, class imbalance~\cite{yang2020rethinkingselfclassimbalance}, etc.
Recently, Lee {\em et al.}~\cite{selfsup} used SSL to improve the performance for supervised classification, by augmenting the original labels using the input transformations. 
In this work, we show that SSL~\cite{selfsup} can be used very effectively for the FSCIL task.\\ \\
\textbf{Stochastic Neural Networks: }
Traditional neural networks cannot model uncertainty well due to their deterministic nature~\cite{blundell2015weightpointestimates}.
Stochastic neural networks~\cite{neal2012bayesian} give robust representations in the form of distributions. 
Subedar {\em et al.}~\cite{subedar2019uncertaintyactivityrecognition} proposed uncertainty aware variational layers for activity recognition. 
Recently, it has been used for person re-identification~\cite{yu2019robustreidentification} and unsupervised domain adaptation~\cite{lu2020stochastic} tasks.
\section{Problem Definition and Notations}
Here, we explain the FSCIL task, which consists of a base task and several incremental stages, and also the notations used in the rest of the paper.
In the base task, the goal is to learn a classifier using large number of labeled samples from several base classes. 
At each incremental step, using a few labeled samples per new class and a single class prototype of the old (previously learnt) classes, the model needs to be updated such that it can classify both the old and the new classes.
Let $\mathcal{D}^{(0)}$ denote the base task which contains large number of annotated data from classes $\mathcal{C}^{(0)}$.
Let the incremental task data be denoted as 
$\{\mathcal{D}^{(1)}, ..,\mathcal{D}^{(t)},.., \mathcal{D}^{(\mathcal{T})}\}$, and the corresponding label spaces be denoted as $\mathcal{C}^{(t)}$, where $t = 1,\hdots, \mathcal{T}$.
Thus, the model will learn a total of $\mathcal{T}$ tasks incrementally and there is no overlap in the label space between the different tasks, i.e. $C^{(t)} \cap C^{(s)} = \phi$; ($t \neq s$).
Once the model has learned on the data $\mathcal{D}^{(t)}$, it has to perform well on all the classes seen so far i.e $\{C^{(0)} \cup C^{(1)} \cup \dots \cup C^{(t)}\}$.
\section{Proposed Method} 
Here, we describe the proposed S3C framework for the FSCIL task.
In many of the initial FSCIL approaches \cite{tao2020few,mazumder2021few,vector_quantization}, the main focus was to develop novel techniques for the incremental step to prevent catastrophic forgetting and overfitting.
Recently, CEC \cite{cec} showed that the base network training has a profound effect on the performance of the incremental tasks.
Using appropriate modifications while learning the base classifier can significantly enhance not only the base class accuracies, but also the performance for the incrementally added classes. 
Even without any fine-tuning during the incremental steps,  CEC reports the state-of-the-art results for FSCIL. 
In the proposed S3C framework, we combine the advantages of both these techniques and propose to not only improve the base classifier training, but also update all the classifiers during the incremental steps.
First, we describe the two main modules of S3C, namely Stochastic Classifier and Self-Supervision and then discuss how to integrate them.  \\ \\
\textbf{Stochastic Classifier:} One of the major challenges in FSCIL is the few number of annotated samples that is available per class at each incremental step. 
This may result in overfitting on the few examples and learning classification boundaries which do not generalize well on the test data. 
Now, we discuss how stochastic classifiers can be used to mitigate this problem.

\begin{figure}[t]
\centering
\begin{subfigure}{0.3\textwidth}
\centering
\begin{center}

\begin{tikzpicture}[scale=0.9]

\fill[gray!1] (0,0) circle (1.5cm);
\draw[black,thick,dashed,opacity=0.7] (0,0) circle (1.5cm);

\draw[black] (270:2.2cm) node {a). Standard Classifier $(\phi)$};

\draw[line width=0.4mm, black!50!green, ->] (0,0) -- (20:1.5cm);
\draw[black!50!green] (20: 1.8cm) node {$\phi_{i}$};
\filldraw[green!80,opacity=0.2] (0,0) -- (20:1.5cm) arc (20:80:1.5cm) -- cycle;

\draw[line width=0.4mm, cyan!90!black, ->] (0,0) -- (75:1.5cm);
\draw[cyan!90!black] (65: 1.9cm) node {$f_{\theta}(\mathbf{x})$};

\draw[line width=0.4mm, blue!50!black, ->] (0,0) -- (140:1.5cm);
\draw[blue!50!black] (140:1.8cm) node {$\phi_{j}$};

\draw[line width=0.2mm,dashdotted, black!10!orange, -] (0,0) -- (80:2.2cm);
\draw[black!10!orange] (80:2.3cm) node {\tiny$m_{ij}$};
\end{tikzpicture}
    
\end{center}
\end{subfigure}
\hspace{1em}
\begin{subfigure}{0.3\textwidth}
\centering
\begin{center}
\begin{tikzpicture}[scale=0.9]

\fill[gray!1] (0,0) circle (1.5cm);
\draw[black,thick,dashed,opacity=0.7] (0,0) circle (1.5cm);

\draw[black] (270:2cm) node {b). 2-Classifiers/Class $(\phi)$};

\draw[line width=0.4mm, black!50!green, ->] (0,0) -- (30:1.5cm);
\draw[line width=0.4mm, black!50!green, ->] (0,0) -- (10:1.5cm);
\draw[black!50!green] (10: 1.8cm) node {$\phi_{i}^{1}$};
\draw[black!50!green] (30: 1.8cm) node {$\phi_{i}^{2}$};
\filldraw[green!80,opacity=0.2] (0,0) -- (10:1.5cm) arc (10:70:1.5cm) -- cycle;

\draw[line width=0.4mm, cyan!90!black, ->] (0,0) -- (75:1.5cm);
\draw[cyan!90!black] (260: 0.25cm) node {$f_{\theta}(\mathbf{x})$};

\draw[line width=0.4mm, blue!50!black, ->] (0,0) -- (130:1.5cm);
\draw[line width=0.4mm, blue!50!black, ->] (0,0) -- (140:1.5cm);
\draw[blue!50!black] (126:1.8cm) node {$\phi_{j}^{1}$};
\draw[blue!50!black] (144:1.8cm) node {$\phi_{j}^{2}$};

\draw[line width=0.2mm,dashdotted, black!10!orange, -] (0,0) -- (70:1.7cm);
\draw[black!10!orange] (65:1.9cm) node {\tiny$m_{ij}^{11}$};

\draw[line width=0.2mm,dashdotted, black!10!orange, -] (0,0) -- (75:2.2cm);
\draw[black!10!orange] (70:2.5cm) node {\tiny$m_{ij}^{12}$};

\draw[line width=0.2mm,dashdotted, black!10!orange, -] (0,0) -- (80:2.2cm);
\draw[black!10!orange] (85:2.4cm) node {\tiny$m_{ij}^{21}$};

\draw[line width=0.2mm,dashdotted, black!10!orange, -] (0,0) -- (85:1.7cm);
\draw[black!10!orange] (90:1.9cm) node {\tiny$m_{ij}^{22}$};
\end{tikzpicture}
    
\end{center}
\end{subfigure}
\hspace{0.9em}
\begin{subfigure}{0.3\textwidth}
\centering
\begin{center}
\begin{tikzpicture}[scale=0.9]

\fill[gray!1] (0,0) circle (1.5cm);
\draw[black,thick,dashed,opacity=0.7] (0,0) circle (1.5cm);
\draw[black!10!orange,opacity=0] (85:2.4cm) node {\tiny$m_{ij}^{21}$};
\draw[black] (275: 2cm) node {c). Stochastic Classifier $(\psi)$};
\filldraw[black!10!orange,opacity=0.5] (0,0) -- (70:2.2cm) arc (70:85:2.2cm) -- cycle;
\draw[black!40!orange] (77.5: 1.8cm) node[rotate=0] {\tiny margin};

\draw[black!50!green] (-10: 1cm) node {$\psi_{i}=\{\mu_{i},\color{magenta!50!pink}\sigma_{i}\color{black!50!green}\}$};
\filldraw[green!80,opacity=0.2] (0,0) -- (10:1.5cm) arc (10:70:1.5cm) -- cycle;
\filldraw[magenta!50!pink,opacity=0.4] (0,0) -- (10:1.5cm) arc (10:30:1.5cm) -- cycle;
\draw[line width=0.4mm, black!50!green, ->] (0,0) -- (20:1.5cm);

\draw[line width=0.4mm, cyan!90!black, ->] (0,0) -- (75:1.5cm);
\draw[cyan!90!black] (100: 1cm) node {$f_{\theta}(\mathbf{x})$};

\filldraw[yellow!90!black] (0,0) -- (130:1.5cm) arc (130:140:1.5cm) -- cycle;
\draw[line width=0.4mm, blue!50!black, ->] (0,0) -- (135:1.5cm);
\draw[blue!50!black] (180:1.2cm) node {$\psi_{j}=\{\mu_{i},\color{yellow!90!black}\sigma_{i}\color{blue!80}\}$};

\end{tikzpicture}
    
\end{center}
\end{subfigure}
\caption{Figure shows the classification boundary between two classes in (a) standard classifier, (b) two-classifiers per class and (c) stochastic classifier. The margin in (c) results in more discriminative classification boundaries.}
\label{diff_normal_stoc}
\end{figure}
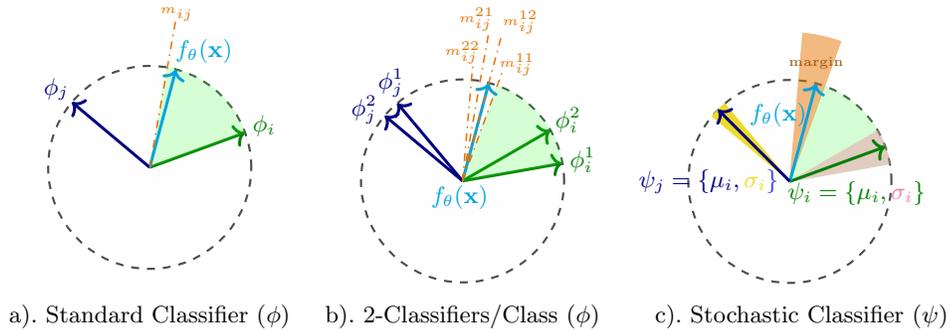

In this work, we use cosine similarity between the features and the classifier weights to compute the class score for that particular feature.
For a given input image $\mathbf{x}$ from class $C_i$, let us denote its feature vector as $f_{\theta}(\mathbf{x})$, where the parameters of the feature extractor $f$ is denoted by $\theta$.
Let the classifier weights corresponding to class $C_i$ be denoted as $\phi_i$.
Then the cosine similarity of the feature with this classifier weight can be computed as $\langle\overline{\phi_{i}},  \overline{f_{\theta}(\mathbf{x})}\rangle$, where $\overline{u}=u/||u||_{2}$ denotes the $l_{2}$ normalized vector.
Fig.~\ref{diff_normal_stoc}(a) shows the normalized feature extractor, and classifier weights for two classes, $C_i$ and $C_j$. The green shaded area denotes the region where $f_{\theta}(\mathbf{x})$ will be correctly classified to class $C_i$, and $m_{ij}$ is the classification boundary between the two classifiers (considering only the upper sector between $\phi_{i}$ and $\phi_{j}$).

Now, instead of a single classifier, let us learn two different classifiers for each class (eg. $\phi^{1}_i$ and $\phi^{2}_i$ for class $C_i$).
In Fig.~\ref{diff_normal_stoc} (b),  $\{m_{ij}^{11},m_{ij}^{12},m_{ij}^{21},m_{ij}^{22}\}$ are the four classification boundaries for four combination of classifiers. 
To ensure that the input data is correctly classified using all the classifiers, the feature embedding $f_{\theta}(\mathbf{x})$ has to move closer to the classifier of its correct class, thus making the samples of a class better clustered and further from samples of other classes. 
But it is difficult to choose (and compute) how many classifiers should be used.
By using a stochastic classifier (Fig.~\ref{diff_normal_stoc} (c)), we can ensure that we have infinite such classifiers around the mean classifier. 
\setlength{\textfloatsep}{8pt}

\begin{figure}[t]
\hspace*{-0.8cm}  
\centering
\includegraphics[scale=0.68]{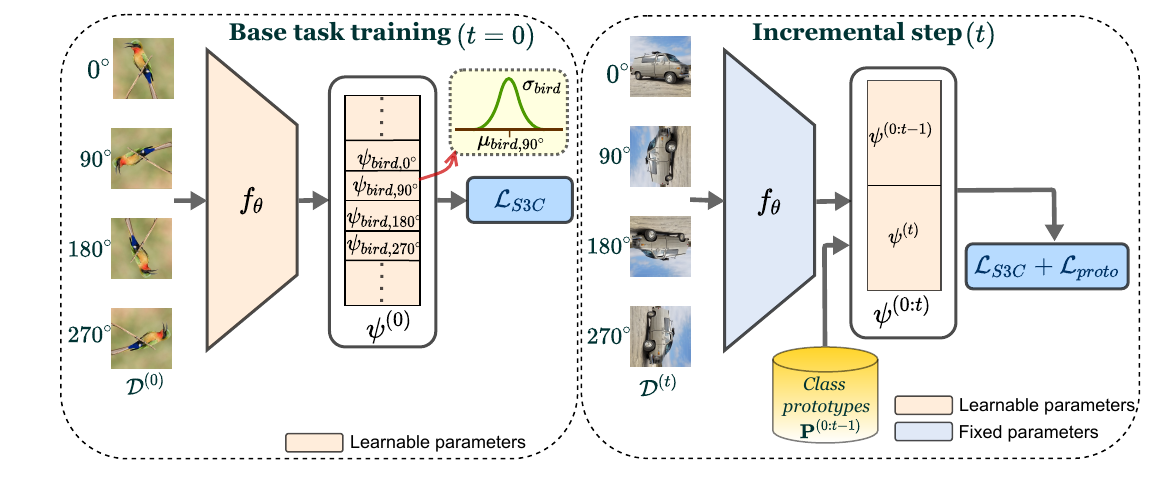}
\caption{Illustration of the proposed S3C framework: Left: Base network training, Right: Training at each incremental step.}
\label{Main}
\end{figure}

Using a stochastic classifier $\psi = \{\mu,\sigma\}$ at the classification head resembles the use of multiple classifiers, where $\mu$ and $\sigma$ denotes the mean and variance of the classifier $\psi$.
For a given input image $\mathbf{x}$, the output score of the stochastic classifiers is proportional to $\langle\overline{\Hat{\mu}},  \overline{f_{\theta}(\mathbf{x})}\rangle$ ($\Hat{\mu} = \mu + \mathcal{N}(0,1) \odot \sigma$), where the classifier is sampled from the distribution. 
This has similarity with feature augmentations which are also commonly used \cite{zhu2021prototype}. 
There are two main advantages of using a stochastic classifier instead of feature augmentations:
(1) Instead of using a fixed variance for the features (which has to be manually calculated), the means and variances used in the proposed framework are automatically learnt in an end-to-end manner.
(2) The means and variances learnt using the base classes also help to initialize the corresponding parameters for the new classes in a semantically meaningful manner as explained later. \\ \\
{\bf Self-supervision:} At the incremental stages, due to presence of few examples from the new classes, in general, most of the FSCIL approaches either fix the feature extractor after learning the base classes~\cite{cec,cheraghian2021semantic} or fine-tune it with a very small learning rate \cite{tao2020few,mazumder2021few,vector_quantization}, so that it does not change significantly.
This reduces catastrophic forgetting as well as overfitting. 
In our work, we fix the feature extractor after learning the base classes and only fine-tune the classifiers. 
To make the base feature extractor generalize well to unseen data, we propose to use self-supervision for the base classifier training as well as during the incremental learning stages.
Since self-supervised training does not use class labels, more generic features can be learnt, which can generalize well to unseen classes. SSL has been used successfully for several tasks \cite{zhai2019s4lselfsemisupervised,hendrycks2019usingselfrobustness,yang2020rethinkingselfclassimbalance,gidaris2019boostingselffewshot,zhu2021prototype}, including the standard class-incremental setting~\cite{zhu2021prototype}. Here, we use the recently proposed SSL approach~\cite{selfsup}, where image augmentations are used to generate artificial labels, which are used to train the classification layer. 
For a given input image $\mathbf{x}$, let the augmented versions be denoted as $\widetilde{\mathbf{x}}_{r}=t_{r}(\mathbf{x})$, where $\{t_{r}\}_{r=1}^{M}$ denotes pre-defined transformations. 
In this work, we use images rotated by $\{0^{\circ},90^{\circ},180^{\circ},270^{\circ}\}$, i.e. (M=$4$) as the augmented images. We show that the feature extractor learnt using  self-supervision performs very well in the incremental stages. 
First, we describe the integrated S3C loss which is used in the training process. \\ \\
\textbf{Construction of S3C loss: }
At task $t$, $C^{(s)}_i$ denotes the $i^{th}$ class in task $s\in\{0,1,..,t\}$.
Then its corresponding stochastic classifier is denoted as $\psi^{(s)}_{i}$ with mean $\mu^{(s)}_{i}$ and variance $\sigma^{(s)}_{i}$. To integrate the stochastic classifiers with self-supervision, for each class, we create four classifier heads corresponding to each of the four rotations as in \cite{selfsup}. 
In this work, we want to jointly predict the class and its rotation $r = \{0^{\circ},90^{\circ},180^{\circ},270^{\circ}\}$, thus we denote the final classifiers as $\psi^{(s)}_{i, r}$, with individual means $(\mu^{(s)}_{i,r})$, but with the same class-wise variance $(\sigma^{(s)}_{i})$.
Since the same data is present in different rotations, we enforce that the classifiers for the same class share the same variances, which reduces the number of parameters to be computed. 
Thus, the joint softmax output of a given sample $\mathbf{x}$ for $C^{(s)}_{i}$ class at $r^{th}$ rotation is given by  
\begin{equation}
    \rho^{(s)}_{ir}(\mathbf{x};\theta,\psi^{(0:t)}) = \frac{exp(\eta \ \langle \ \overline{\Hat{\mu}^{(s)}_{ir}}, \overline{f_{\theta}(\mathbf{x})} \ \rangle)}{\sum\limits_{j=0}^{t} \sum\limits_{k=0}^{|C^{(t)}|}\sum\limits_{l=0}^{M}exp(\eta \ \langle \ \overline{\Hat{\mu}^{(j)}_{kl}}, \overline{f_{\theta}(\mathbf{x})} \ \rangle) }
\end{equation}
Where $\Hat{\mu}^{(j)}_{ir} = \mu^{(j)}_{ir} + \mathcal{N}(0,1) \odot \sigma^{(j)}_{i}$ represents the sampled weight from the stochastic classifier $\psi^{(j)}_{ir}$, $\eta$ is a scaling factor used to control peakiness of the softmax distribution. 
Finally, the S3C training objective for a training sample $\mathbf{x}$ with label $y$ from task $s$ can be written as 

\begin{align}
    \mathcal{L}_{S3C}(\mathbf{x},y;\theta,\psi^{(0:t)}) = -\frac{1}{M} \sum_{r=1}^{M}  \log(\rho^{(s)}_{yr}(\widetilde{\mathbf{x}}_{r};\theta,\psi^{(0:t)}))
    \label{s3c_loss}
\end{align}
This implies that the input image is transformed using the chosen image transformations ($4$ rotations in this work) and the loss is combined for that input. 
Note that the first transformation corresponding to $0^{\circ}$ is the identity transformation (i.e. the original data itself). 
We now describe the base and incremental stage training of the S3C framework (Fig.~\ref{Main}).
\subsection{Base Network Training of S3C} \label{Base_network} 

In FSCIL setting, we assume that we have access to several base classes with sufficient number of annotated data for base training. 
Given the data from the base classes $C^{(0)}$, we use a base network (ResNet20 for CIFAR100 and ResNet18 for CUB200 and miniImageNet) along with a Graph Attention Network inspired by~\cite{cec}\cite{velikovi2017graph}. 
We train the base network, i.e. the feature extractor with parameters $\theta$ and the stochastic classifiers corresponding to the base classes ($\psi^{(0)}$) with S3C objective $\mathcal{L}_{base}=\mathcal{L}_{S3C}(\mathbf{x},y;\theta,\psi^{(0)})$, with the base training data given by $\{\mathbf{x},y\}\in\mathcal{D}^{(0)}$.
The proposed objective improves the performance of the base classes, in addition to that of the new classes that will be encountered in the incremental stages as we will observe in the experimental evaluation.
\subsection{Preparing for the incremental step}
After the base classifier training, the training data of the base classes may not be available any longer.
This may be due to limited storage capacity, privacy issues, etc.
After the first incremental step, we want the unified classifier to perform well on the base as well as on the new classes.
For this, to mitigate catastrophic forgetting of the base classes, their class prototypes are stored as is the common practice~\cite{tao2020few}\cite{cec}.
These stored class prototypes can be treated as class representatives of the base classes and thus can be used for updating the network at the incremental step.
The class prototypes are computed by averaging the training features given by the feature extractor ($f_{\theta}(\cdot)$) for each class.
This is done not only at the end of the base training, but after each incremental step as well, i.e. after incremental step $t$, we store the class prototypes of all the classes that the model has encountered till step $t$.
The class prototype set $\mathbf{P}^{(t)}$ contains the classes prototypes encountered in task $t$.
The class prototype $P_{i}^{(t)}$ after task $t$ for $i^{th}$ class is calculated as 
\begin{align}
    P_{i}^{(t)} = \frac{1}{N^{(t)}_{i}}\sum_{n=1}^{N^{(t)}} \mathbb{I}_{(y_{n}=i)} \ f_{\theta}(\mathbf{x}_{n})
\end{align}
Where $N^{(t)}$ is the number of samples in the dataset $\mathcal{D}^{(t)}$, $N^{(t)}_{i}$ is number of samples in $i^{th}$ class of task $t$, and $\{\mathbf{x}_{n}, y_{n}\}_{n=1}^{N^{(t)}} \in \mathcal{D}^{(t)}$. 
The indicator variable $\mathbb{I}_{(y_{n}=i)}$ will be 1 if the sample belongs to the $i^{th}$ class (i.e. $y_{n}=i$).
Thus, the class prototype set is updated at the end of each task.

\subsection{Incremental Step}
 \label{incremental_step}
Here, we will discuss the training process involved in each incremental step.
As in~\cite{cec} \cite{cheraghian2021semantic}, we propose to freeze the already learnt feature extractor, since the self-supervision has ensured that it will generalize well to previously unseen classes. 
This also helps in mitigating the catastrophic forgetting and over-fitting problems.
In our work, we propose to update the classifiers of the previous as well as the new classes with the stored class-prototypes and the few examples of the new classes.
This will help the model better adapt to the new set of classes. 
Now, we discuss how to initialize the stochastic classifiers for the new classes. \\ 
{\textbf {\em Initialization of the Stochastic Classifiers of the new classes:}} For the new classes, we need to initialize the stochastic classifiers before fine-tuning. The means are initialized with the centroid of the features for that class (calculated using the previous model).
We initialize the variances of the new classes using that of the most semantically similar class from the base set.
Semantic similarity is computed using GloVE embeddings~\cite{pennington2014glove} of the base and new class names.
\\ \\
{\textbf {\em Fine-tuning the classifiers:}}
With this initialization, we fine-tune the classifiers of the new as well as the previous classes using the few labeled examples of the new classes and the stored class-prototypes of the previous classes. Let $q \in \mathbf{P}^{(0:t-1)}$ be a prototype from any old class, then the joint softmax output of the stochastic classifier for $i^{th}$ class and $r^{th}$ rotation (task $s$) is  
\begin{align}
    \zeta^{(s)}_{ir}(q;\psi^{(0:t)}) = \frac{exp(\eta \ \langle \ \overline{\Hat{\mu}^{(s)}_{ir}}, \overline{q} \ \rangle)}{\sum\limits_{j=0}^{t} \sum\limits_{k=0}^{|C^{(t)}|}\sum\limits_{l=0}^{M} exp(\eta \ \langle \ \overline{\Hat{\mu}^{(j)}_{kl}}, \overline{q} \ \rangle) }
\end{align}
\color{black}
For fair comparison with the state-of-the-art approaches, we only store a single class-prototype per class corresponding to the original images (i.e. $0^{\circ}$ rotation). 
Thus for the previous classes, only the parameters of the stochastic classifier corresponding to the $0^{\circ}$ rotation are updated.
To mitigate catastrophic forgetting, we use cross entropy loss based on the class prototypes as
 \begin{align}
   \mathcal{L}_{proto}(q,\check{y},\psi^{(0:t)}) = -\log(\zeta^{(s)}_{\check{y}r}(q;\psi^{(0:t)}))  
 \end{align}
where $\check{y}$ is the class label of the prototype in task $s$. 

For the new classes, very few labeled samples per class is available.
Since the few examples cannot cover the entire distribution, generalization to new classes is quite challenging.
As discussed before, we propose to use stochastic classifiers which mitigates the problem of overfitting and generalizes well to the new classes even with few examples. 
To this end, we calculate a loss as  in equation~(\ref{s3c_loss}) on the new task data using stochastic classifiers. 
Finally, the total loss at each incremental task is given by 
\begin{align}
    \mathcal{L}^{(t)}_{inc} = \lambda_{1} \cdot \mathcal{L}_{proto}(q,\check{y},\psi^{(0:t)}) + \lambda_{2} \cdot \mathcal{L}_{S3C}(\mathbf{x},y;\theta,\psi^{(0:t)}) 
\end{align}
where $\{\mathbf{x},y\}\in \mathcal{D}^{(t)}$ and $t>0$.
$\lambda_{1}, \lambda_{2}$ are hyper-parameters to balance the performance between old and new classes. 
At the end of task $t$, we have the learnt classifiers for all the classes seen so far, namely $\psi^{(0)}, \hdots, \psi^{(t)}$.

\section{Testing Phase}
At inference time, the test image $\mathbf{x}$ can belong to any of the classes seen so far. 
To utilize the learnt classifiers effectively, we generate transformed versions of $\mathbf{x}$ and aggregate all the corresponding scores.
Thus, the aggregate score for the $i^{th}$ class in task $s$ is computed as 
$z^{(s)}_{i}=\frac{1}{M}\sum_{r=1}^{M}\eta \ \langle \ \overline{\mu^{(s)}_{ir}}, \overline{f_{\theta}(\widetilde{\mathbf{x}}_{r})} \rangle$. 
Then the aggregated probability used for predicting the class is given by 
\begin{align}
    P_{agg}(i,s/\mathbf{x},\theta,\psi^{(0:t)}) = \frac{\exp{(z^{(s)}_{i})}}{\sum\limits_{j=0}^{t}\sum\limits_{k=1}^{|C^{(j)}|}\exp{(z^{(j)}_{k})}} 
\end{align}
Thus, the final prediction for the test sample $\mathbf{x}$ is 
\begin{align}\Hat{i},\Hat{s} = arg\max\limits_{i,s}P_{agg}(i,s/\mathbf{x})\end{align} 
which implies that the input $\mathbf{x}$  belongs to $\Hat{i}^{th}$ class of task $\Hat{s}$. 
This aggregation scheme improves the model performance significantly.
\section{Experimental Evaluation}
Here, we describe the extensive experiments performed to evaluate the effectiveness of the proposed S3C framework. 
Starting with a brief introduction of the  datasets, we will discuss the performance of the proposed framework on three standard benchmark datasets.
In addition, we also discuss its effectiveness on two real and challenging scenarios, where (i) the data may be imbalanced at each incremental step and (ii) fewer classes may be available during base training.
We also describe the ablation study to understand the usefulness of each module. \\ \\
\textbf{Datasets Used: } To evaluate the effectiveness of the proposed S3C framework, we perform experiments on three benchmark datasets, namely CIFAR100 \cite{cifar}, miniImageNet \cite{krizhevsky2012imagenet} and CUB200 \cite{cub}. \\
\textbf{CIFAR100}~\cite{cifar} contains $32\times32$ RGB images from $100$ classes, where each class contains $500$ training and $100$ testing images.
We follow the same FSCIL dataset splits as in~\cite{cec}, where the base task is trained with $60$ classes and the remaining $40$ classes is trained in eight incremental tasks in a $5$-$way$ $5$-$shot$ setting. 
Thus, there are a total of $9$ training sessions (i.e., base + $8$ incremental).\\
\textbf{MiniImageNet}~\cite{krizhevsky2012imagenet} is a subset of the ImageNet dataset and contains $100$ classes with images of size $84\times84$. Each class has $600$ images, $500$ for training and 100 for testing. We follow the same task splits as in~\cite{cec}, where $60$ classes are used for base task training and the remaining $40$ classes are learned incrementally in $8$ tasks.
Each task contains $5$ classes with $5$ images per class. \\
\textbf{CUB200}~\cite{cub} is a fine-grained birds dataset with $200$ classes. 
It contains a total of $6000$ images for training and $6000$ images for testing. All the images are resized to $256\times256$ and then cropped to $224\times224$ for training.
We used the the same data splits proposed in~\cite{cec}, where there are $100$ classes in the base task, and each of the $10$ incremental tasks are learned in a $10$-$way$ $5$-$shot$ manner. \\ \\
\textbf{Implementation details: }For fair comparison, we use the same backbone architecture as the previous FSCIL methods~\cite{cec}.
We use ResNet20 for CIFAR100 and ResNet18 for miniImageNet and CUB200 as in~\cite{cec}. 
Inspired by CEC~\cite{cec}, we used the same GAT layer at the feature extractor output for better feature representations. We trained the base network for $200$ epochs with a learning rate of $0.1$ and reduced it to $0.01$ and $0.001$ after $120$ and $160$ epochs for CIFAR100 and miniImageNet datasets. 
For CUB200, the initial learning rate was $0.03$ and was decreased to $0.003$, $0.0003$ after $40$ and $60$ epochs. 
We freeze the backbone network and fine-tune the stochastic classifiers for $100$ epochs with a learning rate of $0.01$ for CIFAR100 and miniImageNet and $0.003$ for CUB200 at each incremental step. 
The base network was trained with a batch size of $128$, and for the newer tasks, we used all the few-shot samples in a  mini-batch for incremental learning. 
All the experiments are run on a single NVIDIA RTX A5000 GPU using PyTorch. We set $\eta=16$, $\lambda_{1}=5$ and $\lambda_{2}=1$ for all our experiments.\\ \\
\textbf{Evaluation protocol: } We evaluate the proposed framework using the following three evaluation metrics as followed in the FSCIL literature: 
(1) First, at the end of each task, we report the \textbf{Top1 accuracy}~\cite{tao2020few,mazumder2021few,cec,vector_quantization} of all the classes seen so far, which is the most commonly used metric;
(2) To be practically useful, the model needs to perform well on all the tasks seen so far (i.e. have a good performance balance between the previous and new tasks). 
To better capture this performance balance, inspired from \cite{xian2017zero}, recent FSCIL works \cite{bhat2021semgifhm_metric,cheraghian2021semantic} propose to use the \textbf{Harmonic Mean} (HM) of the performance of the previous and new classes at the end of each incremental task. 
If $t$ denotes the task id, $t\in\{0,1,...,\mathcal{T}\}$, let $Acc_{n}^{t}$ denote the model accuracy on test data of task $n$ after learning task $t$, where $n\in\{0,1,2,...,t\}$. 
Then at the end of task $t$, to analyze the contribution of base and novel classes in the final accuracy, harmonic mean is calculated between $Acc_{0}^{t}$ and $Acc_{1:t}^{t}$. 
Inspired by CEC, we also report  \textbf{performance dropping rate} $(PD=Acc_{0}^{0}-Acc_{0:\mathcal{T}}^{\mathcal{T}})$ that measures the absolute difference between initial model accuracy after task $0$ and model accuracy at the end of all tasks $\mathcal{T}$.
Here, we report the performance of S3C framework for the standard FSCIL setting on all the three benchmark datasets. 
Note that all the compared approaches have used the same backbone architecture, i.e. ResNet20 for CIFAR100 and ResNet18 for miniImageNet and CUB200 datasets.
As mentioned earlier, most of the FSCIL approaches like TOPIC \cite{tao2020few}, Ft-CNN \cite{tao2020few}, EEIL \cite{EEIL}, iCaRL \cite{rebuffi2017icarl}, UCIR \cite{ucir},
adopted this classifier as it is and proposed different techniques in the incremental stage. 
Thus they have the same base task accuracy as can be observed from the results. 
The current state-of-the-art in FSCIL, CEC~\cite{cec} showed that using the same backbone along with appropriate modifications for learning the base classifier can significantly enhance not only the base class accuracies, but also the performance on the incrementally added classes. 
We combine the advantages of both these techniques, i.e. making the base classifier better (using the same backbone), and at the same time, effectively fine-tuning the stochastic-classifiers in S3C. 
Thus the base accuracy of CEC and the proposed S3C is better than the other approaches.
\captionsetup[figure]{skip=-2pt}
\begin{figure}[t]
\centering
\begin{tikzpicture}

\definecolor{color0}{rgb}{0.545098039215686,0.270588235294118,0.0745098039215686}
\definecolor{color1}{rgb}{0,0.75,0.75}

\begin{groupplot}[group style={group name =GSS CIL experiments all datasets,group size=2 by 1,horizontal sep = 1.5cm,vertical sep = 1.5cm,}, title style={anchor=north, yshift=2.5ex} ]
\nextgroupplot[
tick align=outside,
title={(a) CIFAR100 (ResNet20)},
x grid style={white!80!black},
x label style={at={(axis description cs:0.5,0.02)},anchor=north},
xlabel={tasks},
xmajorgrids,
xmin=-1, xmax=9,
xtick pos=left,
xtick={0,1,2,3,4,5,6,7,8},
xtick style={color=white!15!black},
y grid style={white!80!black},
y label style={at={(axis description cs:0.08,.5)}},
ylabel={top1 accuracy $(\%)$},
ymajorgrids,
ymajorticks=true,
ymin=0, ymax=85,
ytick pos=left,
ytick style={color=white!15!black},
width = 6cm, height = 4cm,
legend to name={CommonLegend},legend style={legend columns=7, legend entries={Ft-CNN,iCaRL,EEIL,UCIR,TOPIC,CEC,S3C (ours)}, legend cell align={left},/tikz/every even column/.append style={column sep=0.5cm},,nodes={scale=0.57, transform shape}}
]
\coordinate (c1) at (rel axis cs:0,1);
\addplot [thick, color0, mark=*,opacity=1, mark size=1, ]
table {
0 64.1
1 36.91
2 15.37
3 9.8
4 6.67
5 3.8
6 3.7
7 3.14
8 2.65
};
\addplot [thick, black, opacity=1,mark=*, mark size=1,]
table {
0 64.1
1 53.28
2 41.69
3 34.13
4 27.93
5 25.06
6 20.41
7 15.48
8 13.73
};
\addplot [thick, yellow, opacity=1,mark=*, mark size=1]
table {
0 64.1
1 53.11
2 43.71
3 35.15
4 28.96
5 24.98
6 21.01
7 17.26
8 15.85
};
\addplot [ thick, lime, mark=*,opacity=1, mark size=1]
table {
0 64.1
1 53.05
2 43.96
3 36.97
4 31.61
5 26.73
6 21.23
7 16.78
8 13.54
};
\addplot [thick, pink!80!black,  mark=*,opacity=1, mark size=1,]
table {
0 64.1
1 55.88
2 47.07
3 45.16
4 40.11
5 36.38
6 33.96
7 31.55
8 29.37
};

\addplot [thick, orange, opacity=1, mark=*, mark size=1]
table {
0 73.07
1 68.88
2 65.26
3 61.19
4 58.09
5 55.57
6 53.22
7 51.34
8 49.14
};

\addplot [thick, cyan!90!black, opacity=1, mark=*, mark size=1]
table {
0 78.63
1 74.50
2 70.68
3 66.42
4 63.82
5 60.52
6 58.64
7 56.88
8 53.96
};


\nextgroupplot[
tick align=outside,
title={(b) miniImageNet (ResNet18)},
x grid style={white!80!black},
x label style={at={(axis description cs:0.5,0.02)},anchor=north},
xlabel={tasks},
xmajorgrids,
xmajorticks=true,
xmin=-1, xmax=9,
xtick pos=left,
xtick={0,1,2,3,4,5,6,7,8},
xtick style={color=white!15!black},
y grid style={white!80!black},
y label style={at={(axis description cs:0.08,.5)}},
ylabel={top1 accuracy $(\%)$},
ymajorgrids,
ymajorticks=true,
ymin=0, ymax=85,
ytick pos=left,
ytick style={color=white!15!black},
width = 6cm, height = 4cm
]
\coordinate (c2) at (rel axis cs:1,2);
\addplot [thick, color0, mark=*,opacity=1, mark size=1, ]
table {
0 61.31
1 27.22
2 16.37
3 6.08
4 2.54
5 1.56
6 1.93
7 2.6
8 1.4
};
\addplot [thick, black, opacity=1,mark=*, mark size=1,]
table {
0 61.31
1 46.32
2 42.94
3 37.63
4 30.49
5 24
6 20.89
7 18.8
8 17.21
};
\addplot [thick, yellow, opacity=1,mark=*, mark size=1]
table {
0 61.31
1 46.58
2 44
3 37.29
4 33.14
5 27.12
6 24.1
7 21.57
8 19.58
};
\addplot [thick, lime, mark=*,opacity=1, mark size=1]
table {
0 61.31
1 47.8
2 39.31
3 31.91
4 25.68
5 21.35
6 18.67
7 17.24
8 14.17
};
\addplot [thick, pink!80!black,  mark=*,opacity=1, mark size=1,]
table {
0 61.31
1 50.09
2 45.17
3 41.16
4 37.48
5 35.52
6 32.19
7 29.46
8 24.42
};

\addplot [thick, orange, opacity=1, mark=*, mark size=1]
table {
0 72.00
1 66.83
2 62.97
3 59.43
4 56.70
5 53.73
6 51.19
7 49.24
8 47.63
};

\addplot [thick, cyan!90!black, opacity=1, mark=*, mark size=1]
table {
0 76.86
1 71.81
2 68.09
3 64.82
4 61.89
5 58.70
6 55.91
7 53.65
8 52.14
};

\end{groupplot}
\coordinate (c3) at ($(c1)!0.47!(c2)$);

\node[above] at (c3 |- current bounding box.north)
{\pgfplotslegendfromname{CommonLegend}};

\end{tikzpicture}
\caption{Comparison of S3C with the state-of-art approaches on CIFAR100 and miniImageNet datasets using the backbone given in the caption.}
\label{cifar_imagenet_results}
\end{figure}
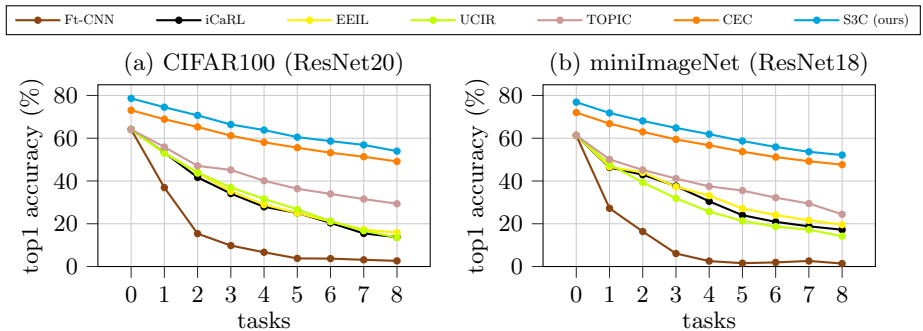
\captionsetup[table]{skip=8pt}
\begin{table}[t]
\centering
\begin{adjustbox}{max width=\linewidth}
\begin{tabular}{llccccccccccc}
\hline \hline
\multirow{2}{*}{Dataset} &\multirow{2}{*}{Method} & \multicolumn{8}{c}{Harmonic Mean $(\%)$ $\uparrow$}                                                                                                                                                                                                                                                \\ \cline{3-11}
                        && \multicolumn{1}{c}{1}     & \multicolumn{1}{c}{2}     & \multicolumn{1}{c}{3}     & \multicolumn{1}{c}{4}     & \multicolumn{1}{c}{5}     & \multicolumn{1}{c}{6}     & \multicolumn{1}{c}{7}     & \multicolumn{1}{c}{8}                                                                                             \\ \hline \hline
                                                                            
\multirow{2}{*}{CIFAR100}       &CEC \cite{cec}                    & \multicolumn{1}{c}{41.57} & \multicolumn{1}{c}{38.75} & \multicolumn{1}{c}{32.36} & \multicolumn{1}{c}{31.53} & \multicolumn{1}{c}{32.55} & \multicolumn{1}{c}{32.40} & \multicolumn{1}{c}{32.25} & \multicolumn{1}{c}{31.27}                                                                                  \\ 
&\textbf{S3C (Ours)}                   & \multicolumn{1}{c}{\textbf{61.60}}      & \multicolumn{1}{c}{\textbf{54.57}}      & \multicolumn{1}{c}{\textbf{48.94}}      & \multicolumn{1}{c}{\textbf{47.60}}      & \multicolumn{1}{c}{\textbf{47.00}}      & \multicolumn{1}{c}{\textbf{46.75}}      & \multicolumn{1}{c}{\textbf{45.96}}      & \multicolumn{1}{c}{\textbf{45.22}}          &                                                                                       \\ \hline
                                                                            
\multirow{2}{*}{miniImageNet}       &CEC \cite{cec}      & \multicolumn{1}{c}{31.68} & \multicolumn{1}{c}{30.86} & \multicolumn{1}{c}{29.52} & \multicolumn{1}{c}{29.01} & \multicolumn{1}{c}{26.75} & \multicolumn{1}{c}{24.46} & \multicolumn{1}{c}{26.14} & \multicolumn{1}{c}{26.24}                                                                                  \\ 
&\textbf{S3C (Ours)}  & \multicolumn{1}{c}{\textbf{35.30}}      & \multicolumn{1}{c}{\textbf{38.18}}      & \multicolumn{1}{c}{\textbf{40.62}}      & \multicolumn{1}{c}{\textbf{38.86}}      & \multicolumn{1}{c}{\textbf{35.02}}      & \multicolumn{1}{c}{\textbf{34.49}}      & \multicolumn{1}{c}{\textbf{36.06}}      & \multicolumn{1}{c}{\textbf{36.20}}          &                                                                                       \\ \hline \hline
\end{tabular}
\end{adjustbox}
\caption{Comparison of S3C with the state-of-the-art CEC in terms of Harmonic Mean on CIFAR100 and miniImageNet datasets. On both datasets S3C outperforms CEC by a considerable margin.}
\label{cifar_imagenet_hm}
\end{table}

\captionsetup[figure]{skip=10pt}
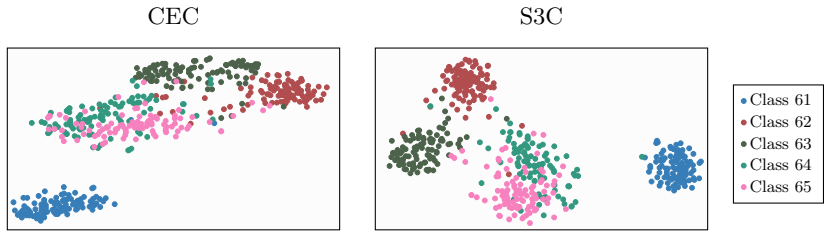
\begin{figure}[t]
\centering
\begin{tikzpicture}
\definecolor{color1}{rgb}{0.215686,0.494118,0.721569}
\definecolor{color2}{rgb}{0.694118,0.301961,0.309804}
\definecolor{color3}{rgb}{0.301961,0.386275,0.290196}
\definecolor{color4}{rgb}{0.2,0.598039215686275,0.5}
\definecolor{color5}{rgb}{0.968627450980392,0.505882352941176,0.749019607843137}

\begin{axis}[
title = {CEC},
ticks=none,
tick pos=left,
xmin=-11.0509518623352, xmax=14.4238747596741,
ymin=-36.3534722328186, ymax=21.8082928657532,
width = 6cm, height = 4cm,
legend style={nodes={scale=0.7, transform shape}, at={(1,0.44)}}, 
        legend image post style={mark=*},
        axis background/.style={fill=gray!2}
]

\addplot [ color1,
  only marks,
 mark = *, mark size=0.9 ,mark options={fill=color1}]
table [x=x, y=y, meta=label]{%
Images/tsne_data/CEC_data_features_tsne_classes_from0to100.dat};
\addplot [color2,
  only marks,
  mark = *, mark size=0.9,mark options={fill=color2}]
table [x=x, y=y, meta=label]{%
Images/tsne_data/CEC_data_features_tsne_classes_from100to200.dat};
\addplot [color3,
  only marks,
  mark = *, mark size=0.9,mark options={fill=color3}]
table [x=x, y=y, meta=label]{%
Images/tsne_data/CEC_data_features_tsne_classes_from200to300.dat};
\addplot [color4,
  only marks,
  mark = *, mark size=0.9,mark options={fill=color4}]
table [x=x, y=y, meta=label]{%
Images/tsne_data/CEC_data_features_tsne_classes_from300to400.dat};
\addplot [color5,
  only marks,
   mark = *, mark size=0.9,mark options={fill=color5}]
table [x=x, y=y, meta=label]{%
Images/tsne_data/CEC_data_features_tsne_classes_from400to500.dat};
\end{axis}



\end{tikzpicture}
\hskip 10pt
\begin{tikzpicture}
\definecolor{color1}{rgb}{0.215686,0.494118,0.721569}
\definecolor{color2}{rgb}{0.694118,0.301961,0.309804}
\definecolor{color3}{rgb}{0.301961,0.386275,0.290196}
\definecolor{color4}{rgb}{0.2,0.598039215686275,0.5}
\definecolor{color5}{rgb}{0.968627450980392,0.505882352941176,0.749019607843137}
\begin{axis}[
title = {S3C},
ticks=none,
tick pos=left,
xmin=-18.0509518623352, xmax=25.4238747596741,
ymin=-12.3534722328186, ymax=14.8082928657532,
width = 6cm, height = 4cm,
legend style={nodes={scale=0.7, transform shape}, at={(1.35,0.8)}}, 
        legend image post style={mark=*},
        axis background/.style={fill=gray!2}
]

\addplot [ color1,
  only marks,
 mark = *, mark size=0.9 ,mark options={fill=color1}]
table [x=x, y=y, meta=label]{%
Images/tsne_data/S3C_data_features_tsne_classes_from0to100.dat};
\addplot [color2,
  only marks,
  mark = *, mark size=0.9,mark options={fill=color2}]
table [x=x, y=y, meta=label]{%
Images/tsne_data/S3C_data_features_tsne_classes_from100to200.dat};
\addplot [color3,
  only marks,
  mark = *, mark size=0.9,mark options={fill=color3}]
table [x=x, y=y, meta=label]{%
Images/tsne_data/S3C_data_features_tsne_classes_from200to300.dat};
\addplot [color4,
  only marks,
  mark = *, mark size=0.9,mark options={fill=color4}]
table [x=x, y=y, meta=label]{%
Images/tsne_data/S3C_data_features_tsne_classes_from300to400.dat};
\addplot [color5,
  only marks,
   mark = *, mark size=0.9,mark options={fill=color5}]
table [x=x, y=y, meta=label]{%
Images/tsne_data/S3C_data_features_tsne_classes_from400to500.dat};
\legend{Class 61,Class 62,Class 63,Class 64,Class 65}
\end{axis} 

\end{tikzpicture}
\caption{Figure shows t-SNE plot (of test samples) from $5$ new classes after task $1$ for CIFAR100 dataset.}
\label{cifar_tsne_session1}
\end{figure}
\captionsetup[table]{skip=8pt}
\begin{table}[b]
\centering
\begin{adjustbox}{max width=\linewidth}
\begin{tabular}{lccccccccccccc}
\hline \hline
\multirow{2}{*}{Method} & \multicolumn{11}{c}{Accuracy  in each session $(\%)$ $\uparrow$}                                                                                                                                                                                                                                                 & \multirow{2}{*}{PD $\downarrow$} & \multirow{2}{*}{\begin{tabular}[c]{@{}c@{}}Our relative\\  improvement\end{tabular}} \\ \cline{2-12}
                        & \multicolumn{1}{c}{0}     & \multicolumn{1}{c}{1}     & \multicolumn{1}{c}{2}     & \multicolumn{1}{c}{3}     & \multicolumn{1}{c}{4}     & \multicolumn{1}{c}{5}     & \multicolumn{1}{c}{6}     & \multicolumn{1}{c}{7}     & \multicolumn{1}{c}{8}     & \multicolumn{1}{c}{9}     & 10    &                     &                                                                                      \\ \hline \hline
Ft-CNN \cite{tao2020few}                 & \multicolumn{1}{c}{68.68} & \multicolumn{1}{c}{43.7}  & \multicolumn{1}{c}{25.05} & \multicolumn{1}{c}{17.72} & \multicolumn{1}{c}{18.08} & \multicolumn{1}{c}{16.95} & \multicolumn{1}{c}{15.1}  & \multicolumn{1}{c}{10.6}  & \multicolumn{1}{c}{8.93}  & \multicolumn{1}{c}{8.93}  & 8.47  & 60.21               &     \textbf{$+$39.83}                                                                                 \\ 
iCaRL \cite{rebuffi2017icarl}                  & \multicolumn{1}{c}{68.68} & \multicolumn{1}{c}{52.65} & \multicolumn{1}{c}{48.61} & \multicolumn{1}{c}{44.16} & \multicolumn{1}{c}{36.62} & \multicolumn{1}{c}{29.52} & \multicolumn{1}{c}{27.83} & \multicolumn{1}{c}{26.26} & \multicolumn{1}{c}{24.01} & \multicolumn{1}{c}{23.89} & 21.16 & 47.52               &    \textbf{$+$26.69}                                                                                  \\ 
EEIL \cite{EEIL}                     & \multicolumn{1}{c}{68.68} & \multicolumn{1}{c}{53.63} & \multicolumn{1}{c}{47.91} & \multicolumn{1}{c}{44.2}  & \multicolumn{1}{c}{36.3}  & \multicolumn{1}{c}{27.46} & \multicolumn{1}{c}{25.93} & \multicolumn{1}{c}{24.7}  & \multicolumn{1}{c}{23.95} & \multicolumn{1}{c}{24.13} & 22.11 & 46.57               &     \textbf{$+$25.74}                                                                                 \\ 
UCIR \cite{ucir}                 & \multicolumn{1}{c}{68.68} & \multicolumn{1}{c}{57.12} & \multicolumn{1}{c}{44.21} & \multicolumn{1}{c}{28.78} & \multicolumn{1}{c}{26.71} & \multicolumn{1}{c}{25.66} & \multicolumn{1}{c}{24.62} & \multicolumn{1}{c}{21.52} & \multicolumn{1}{c}{20.12} & \multicolumn{1}{c}{20.06} & 19.87 & 48.81               &    \textbf{$+$27.98}                                                                                  \\ 
TOPIC \cite{tao2020few}                   & \multicolumn{1}{c}{68.68} & \multicolumn{1}{c}{62.79} & \multicolumn{1}{c}{54.81} & \multicolumn{1}{c}{49.99} & \multicolumn{1}{c}{45.25} & \multicolumn{1}{c}{41.4}  & \multicolumn{1}{c}{38.35} & \multicolumn{1}{c}{35.36} & \multicolumn{1}{c}{32.22} & \multicolumn{1}{c}{28.31} & 26.28 & 42.40               &   \textbf{$+$21.97}                                                                                 \\ 
CEC \cite{cec}                   & \multicolumn{1}{c}{75.85} & \multicolumn{1}{c}{71.94} & \multicolumn{1}{c}{68.50} & \multicolumn{1}{c}{63.50} & \multicolumn{1}{c}{62.43} & \multicolumn{1}{c}{58.27} & \multicolumn{1}{c}{57.73} & \multicolumn{1}{c}{55.81} & \multicolumn{1}{c}{54.83} & \multicolumn{1}{c}{53.52}  & 52.28 & 23.57               &    \textbf{$+$2.74}                                                                               \\ \hline
\textbf{S3C (Ours)}              & \multicolumn{1}{c}{\textbf{80.62}}      & \multicolumn{1}{c}{\textbf{77.55}}      & \multicolumn{1}{c}{\textbf{73.19}}      & \multicolumn{1}{c}{\textbf{68.54}}      & \multicolumn{1}{c}{\textbf{68.05}}      & \multicolumn{1}{c}{\textbf{64.33}}      & \multicolumn{1}{c}{\textbf{63.58}}      & \multicolumn{1}{c}{\textbf{62.07}}      & \multicolumn{1}{c}{\textbf{60.61}}      & \multicolumn{1}{c}{\textbf{59.79}}      &   \textbf{58.95}    &      \textbf{20.83}               &                                                                                      \\ \hline \hline
\end{tabular}
\end{adjustbox}

\caption{Comparison of S3C with other approaches on CUB200 dataset. All the compared results are directly taken from~\cite{cec}.}
\label{cub_acc}
\end{table}
\subsection{Results on standard FS-CIL propocol}
Here, we report the results on the three benchmark datasets.
Fig.~\ref{cifar_imagenet_results} compares the proposed SC3 framework with the state-of-the-art approaches in terms of top1 accuracy on CIFAR100. 
We observe that the modifications while learning the base classifier improves the performance for both CEC and S3C significantly.
At the end of all tasks, S3C achieves a top1 accuracy of $53.96\%$ compared to $49.14\%$ obtained by the state-of-art CEC (relative improvement is $4.82\%$). 
The performance of all the compared approaches are directly taken from~\cite{cec}.
Table~\ref{cifar_imagenet_hm} shows the HM of S3C at the end of each incremental task.
We observe that S3C obtains a relative improvement of $13.95\%$ compared to CEC in terms of HM. 
This shows the effectiveness of S3C in achieving a better balance between the base and new class performance. 
Fig.~\ref{cifar_tsne_session1} shows the t-SNE plot for new classes after task 1, where we observe that the new classes in S3C are relatively well clustered compared to CEC.
In terms of PD, S3C is close to CEC (higher by $0.7\%$), but it outperforms CEC in terms of the other two metrics, namely top1 accuracy and HM.

From Fig.~\ref{cifar_imagenet_results} (right), we observe that S3C achieves $52.14\%$ top1 accuracy on minIimageNet, with a relative improvement of $4.51\%$ over the second best of $47.63\%$ obtained by CEC.
In terms of HM (Table~\ref{cifar_imagenet_hm}) S3C achieves $9.96\%$ relative improvement over CEC. Performance dropping rate (PD) of CEC is slightly lower ($0.35\%$) than S3C.

 We observe from Table~\ref{cub_acc} and Table~\ref{cub_hm} that S3C outperforms CEC by $6.67\%$ and $11.72\%$ respectively in terms of top1 accuracy and HM for CUB200 dataset. 
For this dataset, the proposed S3C has the least performance dropping (PD) rate compared to all the other approaches.

\begin{table}[t]
\centering
\begin{adjustbox}{max width=\linewidth}
\begin{tabular}{lccccccccccccc}
\hline \hline
\multirow{2}{*}{Method} & \multicolumn{10}{c}{Harmonic Mean $(\%)$ $\uparrow$}                                                                                                                                                                                                                                                \\ \cline{2-12}
                            & \multicolumn{1}{c}{1}     & \multicolumn{1}{c}{2}     & \multicolumn{1}{c}{3}     & \multicolumn{1}{c}{4}     & \multicolumn{1}{c}{5}     & \multicolumn{1}{c}{6}     & \multicolumn{1}{c}{7}     & \multicolumn{1}{c}{8}     & \multicolumn{1}{c}{9}     & 10                                                                                   \\ \hline \hline
                                                                            
CEC \cite{cec}                   & \multicolumn{1}{c}{57.63} & \multicolumn{1}{c}{52.83} & \multicolumn{1}{c}{45.08} & \multicolumn{1}{c}{45.97} & \multicolumn{1}{c}{44.44} & \multicolumn{1}{c}{45.63} & \multicolumn{1}{c}{45.10} & \multicolumn{1}{c}{43.76} & \multicolumn{1}{c}{45.77}  & 44.69                                                                                \\ 
\textbf{S3C (Ours)}                   & \multicolumn{1}{c}{\textbf{76.29}}      & \multicolumn{1}{c}{\textbf{65.12}}      & \multicolumn{1}{c}{\textbf{57.30}}      & \multicolumn{1}{c}{\textbf{60.63}}      & \multicolumn{1}{c}{\textbf{56.59}}      & \multicolumn{1}{c}{\textbf{57.79}}      & \multicolumn{1}{c}{\textbf{56.73}}      & \multicolumn{1}{c}{\textbf{55.43}}      & \multicolumn{1}{c}{\textbf{55.48}}      &   \textbf{56.41}    &                                                                                    \\ \hline \hline
\end{tabular}
\end{adjustbox}
\caption{Harmonic mean comparison of S3C with CEC on CUB200 dataset.}
\label{cub_hm}
\end{table}
\subsection{Analysis and Ablation}
\begin{figure}[t]
\centering
\begin{tikzpicture}

\definecolor{color0}{rgb}{0.545098039215686,0.270588235294118,0.0745098039215686}
\definecolor{color1}{rgb}{0,0.75,0.75}

\begin{groupplot}[group style={group name =GSS CIL experiments all datasets,group size=1 by 1,horizontal sep = 1.5cm,vertical sep = 1.5cm,}, title style={anchor=north, yshift=2.5ex} ]
\nextgroupplot[
tick align=outside,title={(a) FSCIL-im},
x grid style={white!80!black},
x label style={at={(axis description cs:0.5,0.02)},anchor=north},
xlabel={tasks},
xmajorgrids,
xmin=-1, xmax=9,
xtick pos=left,
xtick={0,1,2,3,4,5,6,7,8},
xtick style={color=white!15!black},
y grid style={white!80!black},
y label style={at={(axis description cs:0.08,.5)}},
ylabel={Metrics $(\%)$},
ymajorgrids,
ymajorticks=true,
ymin=20, ymax=90,
ytick pos=left,
ytick style={color=white!15!black},
width = 6cm, height = 4cm,
legend to name={CommonLegend},legend style={draw=none,legend columns=1, legend entries={CEC-Acc, S3C-Acc, CEC-HM, S3C-HM}, legend cell align={left},/tikz/every even column/.append style={column sep=0.5cm},nodes={scale=0.55, transform shape}},mark options=solid]
\coordinate (c1) at (rel axis cs:0.80,0.5);
\addplot [thick, orange, mark=*,opacity=1, mark size=1, ]
table {
0 73.03
1 68.69
2 64.74
3 60.26
4 56.57
5 53.63
6 51.22
7 49.33
8 47.13
};
\addplot [thick, cyan!90!black, opacity=1,mark=*, mark size=1,]
table {
0 78.63
1 74.49
2 69.71
3 65.29
4 62.18
5 58.95
6 56.74
7 55.24
8 52.92
};
\addplot [thick, orange, dotted,opacity=1,mark=*, mark size=1, every mark/.append style={solid, fill=orange}]
table {
0 73.03
1 36.27
2 34.24
3 26.64
4 25.64
5 25.93
6 26.20
7 26.27
8 25.24
};
\addplot [thick, cyan!90!black, mark=*,opacity=1, dotted,mark size=1, every mark/.append style={solid, fill=cyan!90!black}]
table {
0 78.63
1 55.05
2 42.95
3 36.87
4 36.35
5 37.37
6 37.39
7 38.49
8 38.94
};
\end{groupplot}

\node at (3.58,1.85)
{\pgfplotslegendfromname{CommonLegend}};
\end{tikzpicture}
\hskip 10pt
\begin{tikzpicture}
\begin{groupplot}[group style={group name =GSS CIL experiments all datasets,group size=1 by 1,horizontal sep = 1.5cm,vertical sep = 1.5cm,}, title style={anchor=north, yshift=2.5ex} ]
\nextgroupplot[
tick align=outside,,title={(b) FSCIL-lb},
x grid style={white!80!black},
x label style={at={(axis description cs:0.5,0.02)},anchor=north},
xlabel={tasks},
xmajorgrids,
xmin=-1, xmax=9,
xtick pos=left,
xtick={0,1,2,3,4,5,6,7,8},
xtick style={color=white!15!black},
y grid style={white!80!black},
y label style={at={(axis description cs:0.08,.5)}},
ylabel={Metrics $(\%)$},
ymajorgrids,
ymajorticks=true,
ymin=20, ymax=90,
ytick pos=left,
ytick style={color=white!15!black},
width = 6cm, height = 4cm,
legend to name={CommonLegend},legend style={draw=none,legend columns=1, legend entries={CEC-Acc, S3C-Acc, CEC-HM, S3C-HM}, legend cell align={left},/tikz/every even column/.append style={column sep=0.5cm},nodes={scale=0.55, transform shape}},mark options=solid]
\coordinate (c1) at (rel axis cs:0.73,1);
\addplot [thick, orange, mark=*,opacity=1, mark size=1, ]
table {
0 78.42
1 71.24
2 64.91
3 59.27
4 55.18
5 52.01
6 49.19
7 46.59
8 44.28
};
\addplot [thick, cyan!90!black, opacity=1,mark=*, mark size=1,]
table {
0 82.09
1 75.75
2 70.23
3 64.03
4 60.93
5 57.63
6 55.55
7 53.27
8 50.17
};
\addplot [thick, orange, dotted,opacity=1,mark=*, mark size=1, every mark/.append style={solid, fill=orange}]
table {
0 78.42
1 33.77
2 31.91
3 27.27
4 27.26
5 27.78
6 28.42
7 27.35
8 27.12
};
\addplot [thick, cyan!90!black, mark=*,opacity=1, dotted,mark size=1, every mark/.append style={solid, fill=cyan!90!black}]
table {
0 82.09
1 63.03
2 55.41
3 49.31
4 47.85
5 47.94
6 48.10
7 46.86
8 45.95
};
\end{groupplot}

\node at (3.58,1.85)
{\pgfplotslegendfromname{CommonLegend}};

\end{tikzpicture}
\caption{Comparison of S3C and CEC in terms of top1 accuracy and harmonic mean for two challenging scenarios, namely (a) FSCIL-im and (b) FSCIL-lb.}
\label{realsitic_scenerios_figure}
\end{figure}
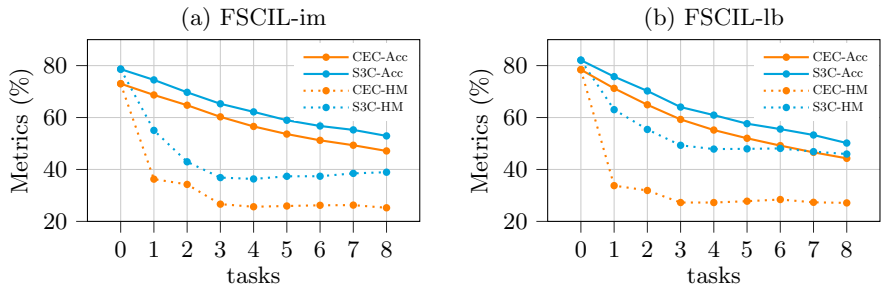

Here, we perform additional experiments and ablation studies on the CIFAR100 dataset to evaluate the effectiveness of the proposed S3C framework. \\ \\
\textbf{Experiments on More Realistic and Challenging Scenarios:}
First, we show the effectiveness of S3C for two realistic scenarios, (i) where there is class imbalance at each incremental task; (ii) where the number of base classes is less. \\ 
\textbf{\em 1. FSCIL-im (imbalance in new classes): } The standard FSCIL setting assumes that equal number of images per new class are available at each incremental task. 
For example, $5$ images for each of the $5$ new classes are available at each incremental task in a $5$-way $5$-shot setting, 
In real-world, number of samples per class can vary, since for some classes, it is easier to collect data compared to others.
Obviously, one can collect more samples from the minority classes, or select a sub-set from the majority classes.
But it is more practical if the algorithm can satisfactorily work without this constraint.

To create the data imbalance, at each incremental step, we consider the number of training samples for the $5$ new classes as
$\{5,4,3,2,1\}$.
Few samples along with the imbalance makes this setting very challenging. 
Fig.~\ref{realsitic_scenerios_figure} (left) shows the top 1 accuracy and HM of S3C and CEC for this scenario without any modification of the algorithms. 
We observe that S3C performs very well for both the metrics, thus showing its effectiveness in handling imbalanced new class data. \\ 
\begin{table}[t]
\centering
\begin{adjustbox}{max width=\linewidth}
\begin{tabular}{ccccc}
\hline \hline
self-supervision &classifier & \begin{tabular}[c]{@{}c@{}}After task $0$ \\  base task accuracy\end{tabular} & \begin{tabular}[c]{@{}c@{}}After task $\mathcal{T}$ \\  top1 accuracy\end{tabular} & \begin{tabular}[c]{@{}c@{}}After task $\mathcal{T}$ \\  harmonic mean\end{tabular}  \\ \hline                                    \xmark   &linear&    74.70  &48.98&26.76   \\  
  \cmark   &linear&    76.14  &53.55& 41.80  \\  
   \cmark   &stochastic&    78.03 &53.96&45.22    
  \\ \hline \hline
\end{tabular}
\end{adjustbox}
\caption{Ablation Study: We observe that both self-supervision and stochastic classifiers help to improve the performance significantly.}
\label{ablation_of_basetask}
\end{table}
\textbf{\em 2. FSCIL-lb (fewer base classes): } The standard FSCIL setting assumes that the number of base classes is quite high, with many annotated samples per class.
Here, we analyze the performance of S3C when the number of base classes is lower.
A similar setting has been explored in \cite{rebuffi2017icarl} for CIL.
The advantage of having lesser number of base classes is that the base learner becomes ready for incremental learning quickly (with fewer classes requiring many annotated samples) and the remaining classes can be learnt incrementally with fewer number of labeled samples per class. 
For the CIFAR100 experiments conducted so far, there were $60$ base and $40$ new classes.
For this experiment, we use only $40$ base classes, and keep the incremental tasks unchanged.
From Fig.~\ref{realsitic_scenerios_figure} (right), we observe that S3C obtains a relative improvement of $5.29\%$ in top1 accuracy ($18.83\%$ in HM) over CEC.
This shows that S3C can start learning incrementally at an early stage of data collection, which makes it more suited for real-world scenarios. \\ \\ 
\textbf{Ablation studies: }
Table~\ref{ablation_of_basetask} shows the effect of self-supervision and type of classifier on CIFAR100 base task accuracy. 
The top 1 accuracy and HM after all the incremental stages are also reported.
We observe that both the modules help in improving the performance of the base and incremental classes.
Though the top 1 accuracy of both linear and stochastic classifiers are close after the incremental stages, there is significant improvement in HM with the stochastic classifier.
This implies that both the modules help in achieving very good performance on the new classes, in addition to retaining the performance on the base, thus achieving a great performance balance between the two.
\section{Conclusions}
In this paper, we proposed a novel S3C framework, which integrates self-supervision with stochastic classifiers seamlessly for the FSCIL task.
We show that this framework not only reduces overfitting on the few labeled samples of the new classes, but also mitigates catastrophic forgetting of the previously learnt classes. Extensive experiments on three benchmark datasets, namely CIFAR100, CUB200 and miniImageNet and additional analysis show that the proposed S3C significantly outperforms the state-of-art approaches.\\
\noindent \textbf{Acknowledgements:} This work is partly supported through a research grant from SERB, Department of Science and Technology, Govt. of India and Google Research, India.
%
%
\bibliographystyle{splncs04}
\bibliography{egbib}

\begin{thebibliography}{10}
\providecommand{\url}[1]{\texttt{#1}}
\providecommand{\urlprefix}{URL }
\providecommand{\doi}[1]{https://doi.org/#1}

\bibitem{akyurek2021subspacereg}
Aky{\"u}rek, A.F., Aky{\"u}rek, E., Wijaya, D., Andreas, J.: Subspace
  regularizers for few-shot class incremental learning. ICLR  (2022)

\bibitem{belouadah2019il2m}
Belouadah, E., Popescu, A.: Il2m: Class incremental learning with dual memory.
  In: ICCV. pp. 583--592 (2019)

\bibitem{belouadah2020scail}
Belouadah, E., Popescu, A.: Scail: Classifier weights scaling for class
  incremental learning. In: WACV. pp. 1266--1275 (2020)

\bibitem{bhat2021semgifhm_metric}
Bhat, S.D., Banerjee, B., Chaudhuri, S.: Semgif: A semantics guided incremental
  few-shot learning framework with generative replay. In: BMVC (2021)

\bibitem{blundell2015weightpointestimates}
Blundell, C., Cornebise, J., Kavukcuoglu, K., Wierstra, D.: Weight uncertainty
  in neural network. In: ICML. pp. 1613--1622 (2015)

\bibitem{caron2018deepimageclustering}
Caron, M., Bojanowski, P., Joulin, A., Douze, M.: Deep clustering for
  unsupervised learning of visual features. In: ECCV. pp. 132--149 (2018)

\bibitem{EEIL}
Castro, F.M., Mar{\'\i}n-Jim{\'e}nez, M.J., Guil, N., Schmid, C., Alahari, K.:
  End-to-end incremental learning. In: ECCV. pp. 233--248 (2018)

\bibitem{vector_quantization}
Chen, K., Lee, C.G.: Incremental few-shot learning via vector quantization in
  deep embedded space. In: ICLR (2020)

\bibitem{cheraghian2021semantic}
Cheraghian, A., Rahman, S., Fang, P., Roy, S.K., Petersson, L., Harandi, M.:
  Semantic-aware knowledge distillation for few-shot class-incremental
  learning. In: CVPR. pp. 2534--2543 (2021)

\bibitem{cheraghian2021mixture}
Cheraghian, A., Rahman, S., Ramasinghe, S., Fang, P., Simon, C., Petersson, L.,
  Harandi, M.: Synthesized feature based few-shot class-incremental learning on
  a mixture of subspaces. In: ICCV. pp. 8661--8670 (2021)

\bibitem{dong2021fewknowledgeretentiongraph}
Dong, S., Hong, X., Tao, X., Chang, X., Wei, X., Gong, Y.: Few-shot
  class-incremental learning via relation knowledge distillation. In: AAAI. pp.
  1255--1263 (2021)

\bibitem{douillard2020podnet}
Douillard, A., Cord, M., Ollion, C., Robert, T., Valle, E.: Podnet: Pooled
  outputs distillation for small-tasks incremental learning. In: ECCV 2020. pp.
  86--102 (2020)

\bibitem{gidaris2019boostingselffewshot}
Gidaris, S., Bursuc, A., Komodakis, N., P{\'e}rez, P., Cord, M.: Boosting
  few-shot visual learning with self-supervision. In: ICCV. pp. 8059--8068
  (2019)

\bibitem{hendrycks2019usingselfrobustness}
Hendrycks, D., Mazeika, M., Kadavath, S., Song, D.: Using self-supervised
  learning can improve model robustness and uncertainty. NeurIPS  \textbf{32}
  (2019)

\bibitem{hinton2015distilling}
Hinton, G., Vinyals, O., Dean, J.: Distilling the knowledge in a neural
  network. NeurIPS Workshop  (2014)

\bibitem{ucir}
Hou, S., Pan, X., Loy, C.C., Wang, Z., Lin, D.: Learning a unified classifier
  incrementally via rebalancing. In: CVPR. pp. 831--839 (2019)

\bibitem{komodakis2018unsupervisedimagerotations}
Komodakis, N., Gidaris, S.: Unsupervised representation learning by predicting
  image rotations. In: ICLR (2018)

\bibitem{cifar}
Krizhevsky, A., Hinton, G.: Learning multiple layers of features from tiny
  images. Master's thesis, Department of Computer Science, University of
  Toronto  (2009)

\bibitem{krizhevsky2012imagenet}
Krizhevsky, A., Sutskever, I., Hinton, G.E.: Imagenet classification with deep
  convolutional neural networks. NeurIPS  \textbf{25},  1097--1105 (2012)

\bibitem{larsson2016learningimagecolorization}
Larsson, G., Maire, M., Shakhnarovich, G.: Learning representations for
  automatic colorization. In: ECCV. pp. 577--593 (2016)

\bibitem{lee2021fewindependentmachanisms}
Lee, E., Huang, C.H., Lee, C.Y.: Few-shot and continual learning with attentive
  independent mechanisms. In: ICCV. pp. 9455--9464 (2021)

\bibitem{selfsup}
Lee, H., Hwang, S.J., Shin, J.: Self-supervised label augmentation via input
  transformations. In: ICML. pp. 5714--5724. PMLR (2020)

\bibitem{li2017learninglwf}
Li, Z., Hoiem, D.: Learning without forgetting. TPAMI  \textbf{40}(12),
  2935--2947 (2017)

\bibitem{lu2020stochastic}
Lu, Z., Yang, Y., Zhu, X., Liu, C., Song, Y.Z., Xiang, T.: Stochastic
  classifiers for unsupervised domain adaptation. In: CVPR. pp. 9111--9120
  (2020)

\bibitem{mazumder2021few}
Mazumder, P., Singh, P., Rai, P.: Few-shot lifelong learning. In: AAAI (2021)

\bibitem{neal2012bayesian}
Neal, R.M.: Bayesian learning for neural networks, vol.~118. Springer Science
  \& Business Media (2012)

\bibitem{noh2015learning}
Noh, H., Hong, S., Han, B.: Learning deconvolution network for semantic
  segmentation. In: ICCV. pp. 1520--1528 (2015)

\bibitem{noroozi2016unsupervisedjigsaw}
Noroozi, M., Favaro, P.: Unsupervised learning of visual representations by
  solving jigsaw puzzles. In: ECCV. pp. 69--84 (2016)

\bibitem{ouyang2016deepid}
Ouyang, W., Zeng, X., Wang, X., Qiu, S., Luo, P., Tian, Y., Li, H., Yang, S.,
  Wang, Z., Li, H., et~al.: Deepid-net: Object detection with deformable part
  based convolutional neural networks. TPAMI  \textbf{39}(7),  1320--1334
  (2016)

\bibitem{pennington2014glove}
Pennington, J., Socher, R., Manning, C.D.: Glove: Global vectors for word
  representation. In: EMNLP. pp. 1532--1543 (2014)

\bibitem{rebuffi2017icarl}
Rebuffi, S.A., Kolesnikov, A., Sperl, G., Lampert, C.H.: icarl: Incremental
  classifier and representation learning. In: CVPR. pp. 2001--2010 (2017)

\bibitem{shi2021overcoming}
Shi, G., Chen, J., Zhang, W., Zhan, L.M., Wu, X.M.: Overcoming catastrophic
  forgetting in incremental few-shot learning by finding flat minima. NeurIPS
  \textbf{34},  6747--6761 (2021)

\bibitem{subedar2019uncertaintyactivityrecognition}
Subedar, M., Krishnan, R., Meyer, P.L., Tickoo, O., Huang, J.:
  Uncertainty-aware audiovisual activity recognition using deep bayesian
  variational inference. In: ICCV. pp. 6301--6310 (2019)

\bibitem{tan2021graphfewshot}
Tan, Z., Ding, K., Guo, R., Liu, H.: Graph few-shot class-incremental learning.
  WSDM  (2022)

\bibitem{tao2020few}
Tao, X., Hong, X., Chang, X., Dong, S., Wei, X., Gong, Y.: Few-shot
  class-incremental learning. In: CVPR. pp. 12183--12192 (2020)

\bibitem{velikovi2017graph}
Veličković, P., Cucurull, G., Casanova, A., Romero, A., Liò, P., Bengio, Y.:
  Graph attention networks. In: ICLR (2017)

\bibitem{cub}
Wah, C., Branson, S., Welinder, P., Perona, P., Belongie, S.: The caltech-ucsd
  birds-200-2011 dataset  (2011),
  \url{http://www.vision.caltech.edu/visipedia/CUB-200.html}

\bibitem{wu2019large}
Wu, Y., Chen, Y., Wang, L., Ye, Y., Liu, Z., Guo, Y., Fu, Y.: Large scale
  incremental learning. In: CVPR. pp. 374--382 (2019)

\bibitem{xian2017zero}
Xian, Y., Schiele, B., Akata, Z.: Zero-shot learning-the good, the bad and the
  ugly. In: CVPR. pp. 4582--4591 (2017)

\bibitem{yang2020rethinkingselfclassimbalance}
Yang, Y., Xu, Z.: Rethinking the value of labels for improving class-imbalanced
  learning. NeurIPS  \textbf{33},  19290--19301 (2020)

\bibitem{yu2020semanticsdc}
Yu, L., Twardowski, B., Liu, X., Herranz, L., Wang, K., Cheng, Y., Jui, S.,
  Weijer, J.v.d.: Semantic drift compensation for class-incremental learning.
  In: CVPR. pp. 6982--6991 (2020)

\bibitem{yu2019robustreidentification}
Yu, T., Li, D., Yang, Y., Hospedales, T.M., Xiang, T.: Robust person
  re-identification by modelling feature uncertainty. In: ICCV. pp. 552--561
  (2019)

\bibitem{zhai2019s4lselfsemisupervised}
Zhai, X., Oliver, A., Kolesnikov, A., Beyer, L.: S4l: Self-supervised
  semi-supervised learning. In: ICCV. pp. 1476--1485 (2019)

\bibitem{cec}
Zhang, C., Song, N., Lin, G., Zheng, Y., Pan, P., Xu, Y.: Few-shot incremental
  learning with continually evolved classifiers. In: CVPR. pp. 12455--12464
  (2021)

\bibitem{zhao2020maintaining}
Zhao, B., Xiao, X., Gan, G., Zhang, B., Xia, S.T.: Maintaining discrimination
  and fairness in class incremental learning. In: CVPR. pp. 13208--13217 (2020)

\bibitem{zhu2021classdual}
Zhu, F., Cheng, Z., Zhang, X.y., Liu, C.l.: Class-incremental learning via dual
  augmentation. NeurIPS  \textbf{34} (2021)

\bibitem{zhu2021prototype}
Zhu, F., Zhang, X.Y., Wang, C., Yin, F., Liu, C.L.: Prototype augmentation and
  self-supervision for incremental learning. In: CVPR. pp. 5871--5880 (2021)

\bibitem{zhu2021selfpromoted}
Zhu, K., Cao, Y., Zhai, W., Cheng, J., Zha, Z.J.: Self-promoted prototype
  refinement for few-shot class-incremental learning. In: CVPR. pp. 6801--6810
  (2021)

\end{thebibliography}
\end{document}